\documentclass[11pt]{article}

\usepackage[preprint]{acl}

\usepackage{times}
\usepackage{latexsym}
\usepackage[T1]{fontenc}
\usepackage[utf8]{inputenc}
\usepackage{microtype}
\usepackage{inconsolata}

\usepackage{amsmath, amssymb}
\usepackage{graphicx}
\usepackage{booktabs}
\usepackage{xcolor}
\usepackage{float}
\usepackage{placeins}

\title{K/V-Cache Interventions Dissociate Representation Alignment from\\
Persona Expression in Decoder-Only Language Models}

\author{Yu Sun \\ \texttt{ysun1@linkedin.com} \And
  Mengyin Lu \\ \texttt{melu@linkedin.com} \And
  Cong Feng \\ \texttt{cofeng@linkedin.com} \AND
  Guangming Lu \\ \texttt{glu@linkedin.com} \And
  Huimin Han \\ \texttt{huhan@linkedin.com}}

\begin{document}
\maketitle

\begin{abstract}

We study K/V-cache interventions---transplanting a target-conditioned K/V trajectory into a source-persona generation---as a structured surface for persona control in decoder-only language models. Across $13$ intervention configurations applied to Llama-3.1-8B for a fixed source$\rightarrow$target persona pair, we report two consistent dissociations between representation-level alignment and behavioral expression, plus a common failure under position-perturbing interventions. First, all layer-band K/V replacement interventions (early, mid, and late) achieve strong local V-space alignment within their replaced layers ($V$-gap $0.91$, $0.89$, $0.84$), but only mid-layer replacement (layers $9$--$20$) combines substantial target-marker expression with a comparatively preserved lexical-diversity profile. Second, interventions that induce comparable V-space alignment (full replacement vs.\ mid-layer replacement, $V$-gap $0.94$ vs.\ $0.89$) produce substantially different lexical-diversity profiles (TTR $0.65$ vs.\ $0.77$). Third, position-perturbing interventions (lag and shuffle of K/V positions) apply distinct positional operations yet uniformly suppress target-persona expression in generated text---a common behavioral failure rather than a representation-behavior dissociation in the strict sense. Partial source--target K/V interpolation produces graded, monotonic behavioral effects over the two tested strengths. These observations indicate that representation-level similarity metrics alone are not sufficient predictors of downstream persona expression in the intervention regimes we study, and locate the K/V cache as a controllable but structurally constrained intervention surface. Because the transplanted trajectory carries the target's own generated token history, we characterize the intervention as trajectory-level transplantation rather than isolated persona-representation injection. A same-token-sequence control, in which the source and target caches decode an identical token sequence and differ only in persona conditioning, reproduces the sign and layer localization of the L$28$ representational shift on the same metric, indicating that the shift is not explained solely by imported token history. These findings characterize the structure of representation--behavior dissociation under K/V intervention in a high-signal setting, rather than establishing universality across models or persona pairs.
\end{abstract}

\section{Introduction}
\label{sec:intro}

Persona control in large language models---broadly, the task of maintaining a specified character, role, or stylistic identity in generated text---is approached through methods that intervene at different points in the autoregressive inference loop: at the input tokens (prompts, in-context examples, prefix tuning), at the model weights (full fine-tuning, low-rank adaptation, RL-based optimization), or at intermediate activations (single- or multi-layer steering vectors injected into the residual stream). The K/V cache is the architectural channel by which past-step computation persists into future steps, yet it has been studied primarily as a target for compression or analysis rather than as a direct intervention surface for behavioral control.

Single-layer activation-channel interventions in our setup exhibit \emph{reset dynamics} (measured in \S\ref{sec:reset-summary}; full methodology in Appendix~\ref{app:activation-channel})---the per-step ratio between controller forcing and model restoring force is $R \approx 1$ with anti-aligned direction on essentially every generation step. This prevents per-step injections from accumulating into a persistent behavioral shift and motivates investigating K/V interventions, which by architectural design persist across generation steps. It is plausible that K/V-level interventions provide a controllable surface that single-layer activation methods do not; it is equally plausible that the structural constraints of attention---layer-by-layer K/V dependencies, position-sensitive routing, and cross-layer manifold consistency---introduce their own limits on what kinds of K/V modifications produce coherent behavior.

This paper investigates the K/V cache as a persona-control intervention surface. Each configuration transplants a target-conditioned reference K/V trajectory (captured from an independent target-persona rollout) into a source-persona generation; we therefore treat the intervention as trajectory-level transplantation rather than isolated persona-representation injection. We construct a battery of $13$ intervention configurations spanning full replacement (v1\_full), partial layer-band replacement (v2 across three coarse and three fine layer bands), position-perturbing operations (v3 shuffle and lag), and source--target interpolation (v4 at two interpolation strengths). For each configuration we measure three complementary signals: text-level persona-marker density, L$28$ hidden-state V$^\star$-projection (a per-axis-rescaled distance to per-persona centroids learned from prior work), and K/V cosine alignment with paired source-prompt and target-prompt reference runs.

\paragraph{Findings.}
We observe three patterns in this battery, each supported by at least two of the three measurement families:

\begin{itemize}
\item \textbf{Mid-layer concentration.} K/V replacement at layers $9$--$20$ produces both target-aligned representations (V-gap $0.89$) and target-marker transfer with a preserved lexical-diversity profile (target marker density $15.5$, TTR $0.77$). Replacements at early ($1$--$8$) or late ($21$--$31$) bands produce high V-alignment within their respective replaced layers (V-gap $0.91$ and $0.84$), but this alignment does not transfer to other layers and does not produce target-marker transfer.
\item \textbf{Alignment does not determine behavior.} Full K/V replacement (v1\_full) achieves the highest V-alignment ($0.94$) and the highest target marker density ($24.8$), but at substantially higher lexical repetition (TTR $0.65$). Mid-layer replacement (v2\_L9-20) achieves nearly equal V-alignment ($0.89$) with a preserved lexical-diversity profile (TTR $0.77$). Comparable representation-level alignment thus corresponds to qualitatively different behavioral outcomes.
\item \textbf{Distinct perturbations yield a common behavioral failure.} Position-perturbing interventions (lag and shuffle of K/V positions) apply qualitatively different operations to the source-prompt K/V cache, yet neither injects target-persona representation and both yield minimal target-persona text expression (target marker density $\leq 0.5$). Here representation and behavior agree---neither shows target transfer---a common failure mode rather than a representation-behavior dissociation in the strict sense.
\end{itemize}

Taken together, these results indicate that representation-level similarity metrics alone---whether measured in K/V space or via hidden-state projection onto a persona subspace---do not sufficiently predict downstream persona expression in the intervention regimes we study.

\paragraph{Contribution.}
We document an empirical regularity: a multi-channel dissociation between representation-level alignment and behavioral expression that holds across $13$ K/V intervention configurations and two complementary representation-space measurements on Llama-3.1-8B with a single source$\rightarrow$target persona pair. The contribution of this work is the characterization of the K/V intervention surface itself. The mid-layer concentration we identify is a design implication of that characterization rather than a benchmark-optimized steering method.

\section{Related Work}
\label{sec:related}

\paragraph{Activation, token, and weight channels.} Single-layer and multi-layer activation injection methods include CAA \citep{panickssery2024caa}, ActAdd \citep{turner2023actadd}, RepE \citep{zou2023repe}, ITI \citep{li2023iti}, function vectors \citep{todd2023function}, and persona vectors \citep{anthropic2025personavectors}. Token-channel methods --- prompting, in-context demonstrations, soft prompts \citep{lester2021softprompt}, and prefix tuning \citep{li2021prefix} --- operate by introducing tokens that the model's causal attention re-reads each step. Weight-channel methods --- fine-tuning, LoRA, RLHF, DPO, and model-editing methods such as ROME \citep{meng2022rome} and MEMIT \citep{meng2023memit} --- modify the parameters of the model itself. We establish in Appendix~\ref{app:activation-channel} that single-layer activation injection in our setup is subject to per-step reset dynamics; the K/V-level constraints we document in this paper are distinct from this activation-channel constraint and from the token- and weight-channel pathways.

\paragraph{K/V-cache analysis and intervention.} Prior work on K/V cache has studied compression \citep{zhang2024h2o}, attention-head specialization \citep{olsson2022induction}, and prefix-tuning-style learned cache prefixes. Direct intervention on the K/V cache as a persona-control surface, with systematic ablation across layer bands, position perturbations, and interpolation strengths, has not been reported in the persona-control literature to our knowledge. Recent work in video character generation \citep{lpm2026} achieves long-horizon identity stability through a combination of multi-reference token-channel conditioning and weight-channel distillation, consistent with our observation that long-horizon persona stability is achievable via these channels.

\paragraph{Representation--behavior dissociation.} The general phenomenon that representation similarity metrics may fail to predict behavioral outcomes has been reported in interpretability literature, e.g.\ in the context of probing classifiers and representation engineering \citep{belrose2023eliciting}. Our observation extends this in two specific directions: (i) we identify dissociation in both directions (similar alignment with different behavior; different misalignment with similar behavior), and (ii) we corroborate across two complementary representation spaces (hidden-state projection and K/V cosine) under the same intervention battery.

\paragraph{Concurrent work.} Three recent studies share our representation-versus-behavior framing in adjacent settings. \citet{jiang2026behavioral} document an ``audit gap'': safety-aligned models that match their base on every static behavioral audit yet give way to small latent-state perturbations, a representation--behavior decoupling established through latent perturbation and fine-tuning in the safety domain rather than through K/V-cache intervention. \citet{kang2026prompt} identify ``K/V-cache contamination''---steered token states stored and reused across steps---as a failure mode of residual-stream activation steering in multi-turn dialogue, and propose a gated attention-delta method (GCAD) to restore coherence; we instead intervene on the K/V cache directly and characterize the resulting representation--behavior structure across a static intervention battery, rather than proposing a control method. \citet{baez2026dissociating} dissociate sycophancy representations into factual and opinion subtypes via linear probes and steering vectors, a precedent for representation-subtype dissociation distinct from our representation--behavior focus.

\section{Methods}
\label{sec:methods}

\subsection{Setup}
\label{sec:setup}

We use Llama-3.1-8B-Instruct \citep{llama31} at bfloat16 precision on a single H100 GPU ($32$ decoder layers, $32$ query heads, $8$ K/V heads with grouped-query attention, head dimension $128$). We focus on a single source $\rightarrow$ target persona pair from a 30-persona corpus introduced in our prior work: \texttt{health\_nurse\_dry} (source ``Nurse Reyes'': $30$-year ER veteran, terse and deadpan) $\rightarrow$ \texttt{tech\_novice\_anxious} (target ``Priya'': coding-bootcamp graduate, anxious and apologetic; full system prompts in Appendix~\ref{app:personas}). The two personas are far apart in V$^\star$-space ($\|c_{\text{tgt}} - c_{\text{src}}\| = 22.5$ raw, $7.2$ in per-axis-rescaled units). The pair was selected as a high-separation setting ($7.2$ vs.\ typical $\sim 3$ across the corpus), which increases signal-to-noise for resolving layer-localized intervention structure; whether the same concentration profile holds at smaller separations remains open. For each intervention configuration we generate $n = 10$ samples at temperature $0.7$ with \texttt{max\_new\_tokens} $= 50$, conditioned on a single seed user message (\texttt{casual\_hobby}: ``\emph{My sourdough starter died. What did I do wrong?}''). The V$^\star$ subspace (10-dim, supervised, at L$28$) and per-persona centroids are reused from prior work without modification.%

\subsection{Intervention battery}
\label{sec:interventions}

We construct $13$ K/V intervention configurations covering four categories.%

\paragraph{v1\_full: full K/V replacement.} We first run a reference forward pass with the target-persona system prompt and the same seed user message, capturing the K/V cache trajectory $\mathcal{R}^{\text{tgt}} = \{(K^{\text{tgt}}_t, V^{\text{tgt}}_t)\}_{t=0}^{T-1}$. During the manipulated run with the source-persona system prompt, after each forward pass we overwrite the entire \texttt{past\_key\_values} state with the target reference at the corresponding step. Because the reference is an independent target-persona rollout, $\mathcal{R}^{\text{tgt}}$ encodes the target's own generated token history; the intervention therefore transplants a target-conditioned trajectory, and we do not isolate the contribution of that generated history from persona conditioning (\S\ref{sec:limitations}).

\paragraph{v2\_partial: layer-band K/V replacement.} Identical to v1\_full but overwriting only at a subset of layers $L \subset \{0, \ldots, 31\}$. We test three coarse bands (L1--8, L9--20, L21--31) and three fine sub-bands within the mid range (L9--12, L13--16, L17--20).

\paragraph{v3 position-perturbing: shuffle and lag.} Structural perturbations to the source-prompt K/V cache without injecting target K/V. After source-prompt prefill the per-layer K/V (along the sequence-position axis) is either randomly permuted (shuffle) or cyclically shifted by $\ell \in \{1, 3, 5\}$ positions (lag).

\paragraph{v4\_interp: source--target K/V interpolation.} Identical to v2\_L9--20 but linearly interpolating each manipulated layer's K/V with the target reference at each step:
\begin{equation}
K^{\text{run}}_{t,l} \leftarrow \alpha \cdot K^{\text{tgt}}_{t,l} + (1 - \alpha) \cdot K^{\text{src}}_{t,l},
\end{equation}
with $l \in L_{9-20}$ and analogously for $V$; we test $\alpha \in \{0.25, 0.5\}$.

\paragraph{v5\_tf: same-token-sequence target-forward control.} The v1/v2 interventions transplant a target-conditioned reference trajectory that carries the target's own generated token history, so they do not separate persona conditioning from imported token history (\S\ref{sec:limitations}). To isolate the persona-conditioning contribution we add a control that removes the token-history difference by construction. We run two K/V caches in lockstep over a \emph{single shared token sequence}---identical tokens at every generation step---differing only in the persona system prompt (source vs.\ target). At each step we overwrite the source-prompt run's K/V (over a layer band $L$) with the target-prompt run's K/V computed \emph{on the same token}, then read the L$28$ hidden state from the manipulated run (the read hook is overwritten per forward pass, so the target-prompt forward is executed first and the manipulated source-prompt forward last, ensuring the recorded hidden state reflects the manipulated run). Because both runs decode the identical realized token sequence, any residual target-ward shift cannot arise from the target having generated a \emph{different} token history. We run four bands---early (L1--8), mid (L9--20), late (L21--31), and full (all $32$ layers)---at $n = 10$ with matched generation parameters. We term this a \emph{same-token-sequence target-forward} control; because the shared token sequence is post-treatment (\S\ref{sec:limitations}), it isolates whether the intervention still moves the internal state \emph{given the same realized tokens}, rather than constituting a fully independent persona counterfactual.

\subsection{K/V dump pipeline}
\label{sec:kv-dump}

We instrument each generation pass to record the last-position K and V tensors at every layer and step, $K_{t,l}, V_{t,l} \in \mathbb{R}^{H \times d}$ with $H = 8$ K/V heads and $d = 128$. \emph{Full head dimension is preserved without averaging.} Per-step logits are recorded at fp$16$ for downstream analyses. For each configuration we additionally record two paired reference K/V trajectories: a source-baseline run (no manipulation) and a target-baseline run. K/V dumps are taken on the first sample only of each configuration's $10$-sample cell (storage cost); this n=1 sampling is a noted limitation (\S\ref{sec:limitations}).

\subsection{Analysis pipeline}
\label{sec:analysis}

\paragraph{Text-density measurement.} We hand-curate two persona-marker phrase sets ($19$ target markers including \texttt{priya, bootcamp, um, uh, oh no, i'm so sorry}; $15$ source markers including \texttt{patient, antibiotics, dosage, 30-year, overfed}; full lists in Appendix~\ref{app:personas}). Marker density is reported as substring matches per $100$ words of generated text, averaged across $10$ samples per configuration. A word-boundary matcher with nested-marker deduplication leaves the cross-condition ordering and the dissociation pattern unchanged (Appendix~\ref{app:marker-sensitivity}); it slightly lowers the mid-band rate (nested \texttt{sorry}/\texttt{i'm so sorry}) and reduces the small apparent transfer in the early and position-perturbing conditions to zero. We report type-token ratio (TTR) not as a holistic quality metric, but as an operational detector of repetition-collapse under aggressive cache perturbation (e.g., the ``three months, three months, three months'' pattern at v1\_full). Two source-marker phrases (\texttt{starter}, \texttt{dead}) overlap with the seed user message vocabulary; we therefore treat source-marker density as a topic-adherence (on-topic fluency) signal and anchor persona-identity claims on target-marker density (see also \S\ref{sec:limitations}).

\paragraph{Hidden-state V$^\star$-projection.} For each generated sample we project the L$28$ residual-stream activation to V$^\star$-space:
\begin{equation}
\tilde{h} = V^\star \cdot (h - \mu) / \sigma,
\end{equation}
where $V^\star \in \mathbb{R}^{10 \times 4096}$ is the supervised subspace basis and $\mu, \sigma$ are standardization parameters. We compute the per-axis-rescaled V$^\star$-distance gap between sample and source/target centroids:
\begin{equation}
\Delta_{\text{V*-gap}} = \| \tilde{h} - c_{\text{src}} \|_{\sigma_{V^\star}} - \| \tilde{h} - c_{\text{tgt}} \|_{\sigma_{V^\star}},
\end{equation}
with $\| x \|_{\sigma} = \sqrt{\sum_i (x_i/\sigma_i)^2}$ and $\sigma_{V^\star}$ the per-axis standard deviation across the $30$ persona centroids. Per-axis rescaling corrects for the $3.2\times$ V$^\star$ anisotropy; raw Euclidean V$^\star$-gap (transparency check) is in Appendix~\ref{app:raw-vstar}.

\paragraph{K/V cosine alignment.} For each manipulated run and each paired reference, we compute per-(layer, step, head) cosine similarity between the manipulated K (resp.\ V) tensor and the reference K (resp.\ V) tensor:
\begin{equation}
\cos_{l,t,h}(\text{run}, \text{ref}) = \frac{\langle X^{\text{run}}_{l,t,h}, X^{\text{ref}}_{l,t,h} \rangle}{\|X^{\text{run}}_{l,t,h}\| \|X^{\text{ref}}_{l,t,h}\| + \epsilon},
\end{equation}
for $X \in \{K, V\}$, with $\epsilon = 10^{-8}$. The K/V cosine gap is $\text{gap}^X = \cos^X(\text{run}, \text{target}) - \cos^X(\text{run}, \text{source})$. We aggregate by mean over step and head for layer-level summaries.

\paragraph{Generation parameters.} Persona pair fixed at \texttt{health\_nurse\_dry} $\rightarrow$ \texttt{tech\_novice\_anxious}; seed topic fixed at \texttt{casual\_hobby}; $n = 10$ samples per cell; \texttt{temperature} $= 0.7$; \texttt{max\_new\_tokens} $= 50$; no activation-level steering applied. Code, configurations, and analysis scripts will be released at \texttt{[anonymized]}.

\section{Results}
\label{sec:results}

\subsection{Reset dynamics motivate the K/V investigation}
\label{sec:reset-summary}

We first verify in our setup that single-layer activation-channel injection at L$24$/L$28$ does not produce persistent behavioral shift under autoregressive inference. Across $20$ closed-loop runs ($4$ controller cells $\times\, n = 5$ samples; full methodology in Appendix~\ref{app:closed-loop-setup}), we measure the per-step \emph{reset ratio} $R_t = \|\Delta_{\text{model}}\| / \|\Delta_{\text{forced}}\|$ between the model's restoring response over the next forward pass and the controller's injection at the hook layer, together with the per-step alignment $\cos(\Delta_{\text{model}}, e_t)$ with the target direction.

\begin{table}[!t]
\centering
\small
\begin{tabular}{llccc}
\toprule
$(\ell, \text{Ctrl})$ & $\alpha_{\max}$ & $\|\Delta_{\text{forced}}\|$ & $R$ & $\cos(\Delta_{\text{model}}, e_t)$ \\
\midrule
L$28$, A & $3$ & $17.8 \pm 0.9$  & $1.00 \pm 0.01$ & $-0.90 \pm 0.01$ \\
L$28$, A & $8$ & $35.5 \pm 2.6$  & $0.98 \pm 0.00$ & $-0.93 \pm 0.00$ \\
L$24$, A & $8$ & $21.8 \pm 2.3$  & $0.99 \pm 0.01$ & $-0.92 \pm 0.00$ \\
L$28$, B & $4$ & $4.0 \pm 0.0$   & $1.64 \pm 0.09$ & $-0.40 \pm 0.05$ \\
\bottomrule
\end{tabular}
\caption{Closed-loop V$^\star$ state-feedback: per-step force decomposition. Values are mean $\pm$ s.d.\ across $n = 5$ samples per cell. $R$ is the median per-step $R_t$ within each sample. The fraction of steps where $\cos < 0$ is $100\%$ for the first three rows and $84\%$ for the controller-B row. Per-step temporal stability and additional controller details are in Appendix~\ref{app:activation-channel}.}
\label{tab:c1-force}
\end{table}

Under the magnitude-adaptive controller (A) at L$28$ and L$24$, the per-step reset ratio is $R \in [0.98, 1.00]$ across the three (layer, $\alpha_{\max}$) cells with sample-level range $[0.98, 1.02]$ across $15$ samples; the model's per-step displacement is anti-aligned with the target direction at $\cos \in [-0.93, -0.90]$ on $100\%$ of generation steps. Controller injections of $17.8$, $21.8$, and $35.5$ V$^\star$-units per step are matched in magnitude and approximately opposite in V$^\star$-direction by the model's intrinsic dynamics within the same step, and do not accumulate into a persistent persona shift. The linear-with-clip controller (B) at $\alpha_{\max} = 4$ produces weaker per-step magnitude ($\|\Delta_{\text{forced}}\| = 4.0$) and shares the qualitative anti-alignment finding ($84\%$ pull-away). Removing token-channel context ($\texttt{window} = 0$) preserves $R \approx 1$ but collapses coherent generation entirely (Appendix~\ref{app:context-window}), separating per-step controller authority from text-level coherence.

We refer to this regularity as \emph{reset dynamics}: under autoregressive inference, single-layer activation injection is restored at the per-step time-scale rather than accumulating into a persistent state shift. Our measurements are at L$24$ and L$28$, both in the late half of the $32$-layer model; whether $R \approx 1$ holds at early or mid hook layers is an open question. The K/V cache, by architectural design, is the channel through which past-step computation \emph{does} persist into future steps; we now investigate K/V interventions as an alternative control surface.

\subsection{Text-level marker density and lexical diversity}
\label{sec:results-text}

\begin{figure}[!t]
\centering
\includegraphics[width=\linewidth]{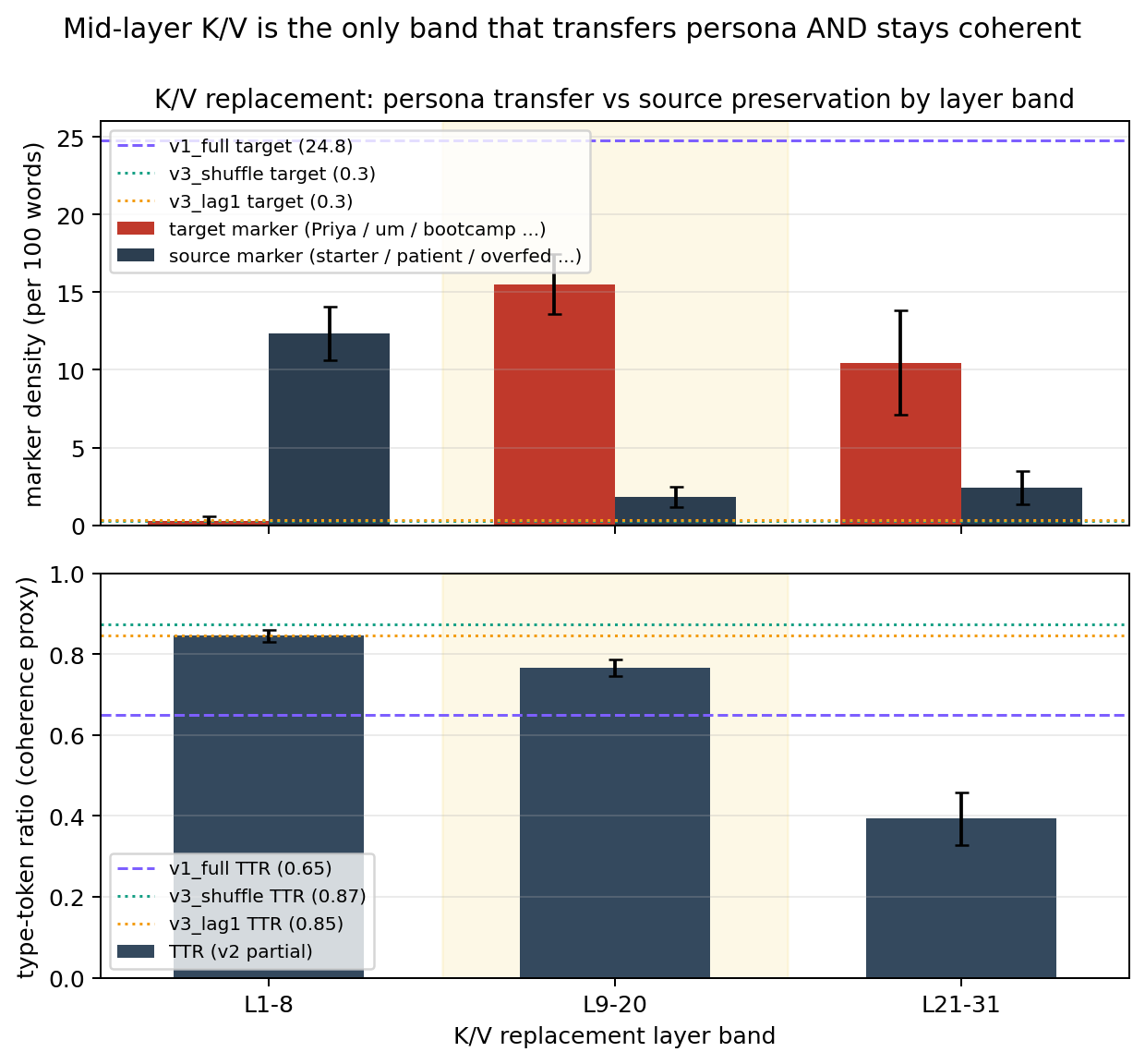}
\caption{Persona-marker density (top) and lexical diversity / TTR (bottom) by K/V intervention layer band. Mid-layer (L9--20, shaded) is the unique band combining target marker density with high TTR. Anchor lines: full replacement (v1, dashed); position perturbations (v3 shuffle and lag, dotted).}
\label{fig:text-density}
\end{figure}

Figure~\ref{fig:text-density} shows target and source persona-marker density per $100$ generated words across the four primary K/V intervention categories. Mid-layer replacement (v2\_L9--20) is the only configuration that produces simultaneous high target marker density ($15.5$) and a preserved lexical-diversity profile (TTR $0.77$). Full replacement (v1\_full) achieves higher target marker density ($24.8$) but with higher lexical repetition (TTR $0.65$): manual inspection reveals repeated phrases such as ``three months, three months, three months'' and ``oh, oh, oh'' indicating output collapse despite successful target-direction representation shift. Late-layer replacement (v2\_L21--31) achieves substantial target marker density ($10.5$) but shows the lowest lexical diversity in the battery (TTR $0.39$). Early-layer replacement (v2\_L1--8) and the position-perturbing v3 conditions all produce no target markers; generated text remains fluent and on-topic for the seed prompt, with topical-vocabulary density $\geq 8$ per $100$ words, consistent with the un-intervened source-prompt regime.%

\subsection{Hidden-state V$^\star$-projection confirms mid-layer concentration}
\label{sec:results-hidden}

\begin{figure}[!t]
\centering
\includegraphics[width=0.95\linewidth]{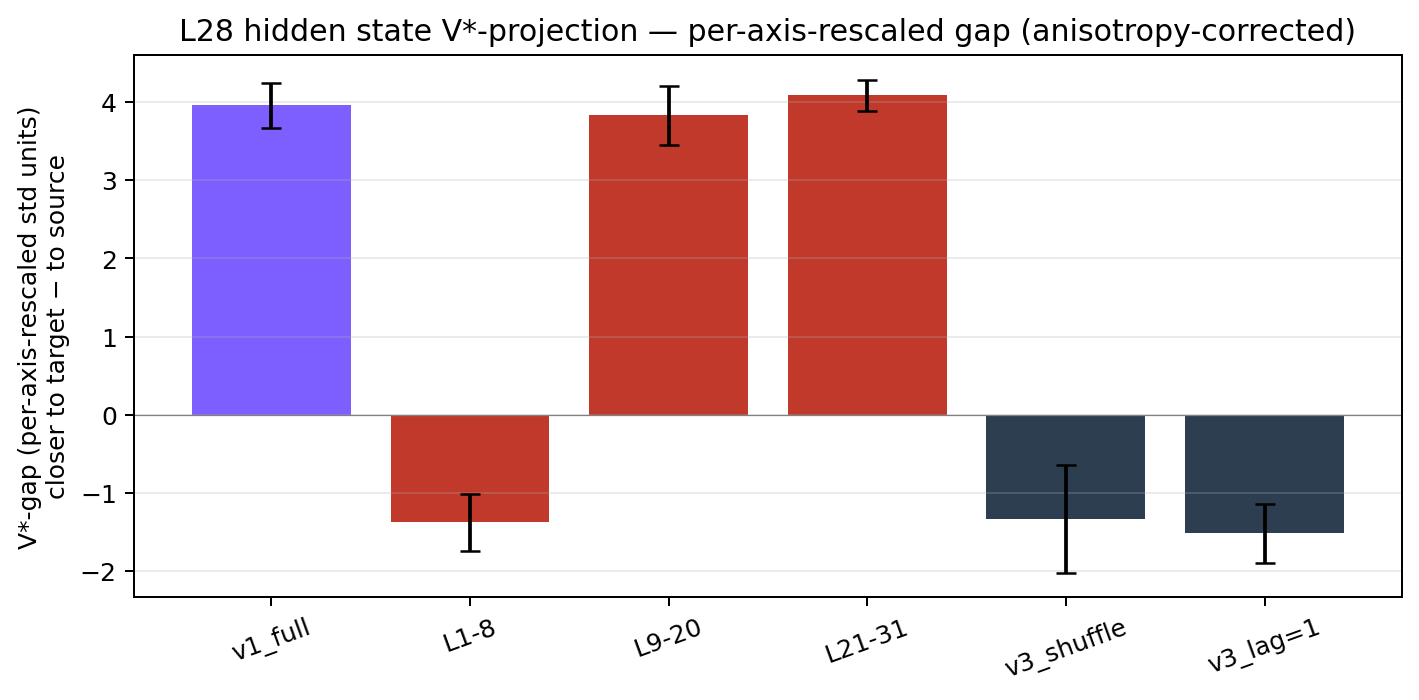}
\caption{L$28$ hidden-state V$^\star$-projection: per-axis-rescaled gap $\| \tilde{h} - c_{\text{src}} \|_{\sigma} - \| \tilde{h} - c_{\text{tgt}} \|_{\sigma}$ (mean $\pm$ s.e.\ across $10$ samples). Full replacement (v1\_full), mid-layer (v2\_L9--20), and late-layer (v2\_L21--31) reach comparable persona-aligned displacement yet differ sharply in lexical diversity.}
\label{fig:hidden-gap}
\end{figure}

Figure~\ref{fig:hidden-gap} shows the per-axis-rescaled V$^\star$-gap for the six conditions on which we compute the L$28$ hidden-state projection (the three high-V-alignment conditions---full, mid-layer, and late-layer replacement---together with early-layer replacement and the two position perturbations); the full $13$-condition battery is characterized through text-marker density and K/V cosine (Table~\ref{tab:full-data}). Across these six conditions, the sign of the V$^\star$-gap agrees with the text-gap, indicating that K/V interventions induce consistent shifts in both hidden-state V$^\star$-projection and text persona-marker density. Three conditions reach comparably high V$^\star$-alignment---full replacement (v1\_full, $+3.96 \pm 0.29$), mid-layer (v2\_L9--20, $+3.83 \pm 0.38$), and late-layer (v2\_L21--31, $+4.09 \pm 0.20$), all within $\sim\!0.3$ rescaled units of one another (per-axis-rescaled units, mean $\pm$ s.e., $n = 10$)---yet these same three conditions span the full range of lexical-diversity values (TTR $0.65$, $0.77$, and $0.39$ respectively). Comparable hidden-state persona alignment thus does not predict the lexical-diversity profile of persona expression. Early-layer replacement (v2\_L1--8, $-1.38 \pm 0.36$) and the v3 position perturbations (shuffle $-1.33 \pm 0.69$; lag$=1$ $-1.52 \pm 0.38$) show negative V$^\star$-gap, consistent with their failure to inject target-direction representation.

\subsection{Same-token-sequence control: the shift is not solely token history}
\label{sec:results-tfcontrol}

The V$^\star$-gap above is measured on interventions (v1/v2) that transplant a target-conditioned trajectory, which carries the target's own generated token history; the resulting shift therefore admits a token-history explanation. The same-token-sequence target-forward control (\S\ref{sec:interventions}) removes that difference by construction: source and target caches decode the \emph{identical} realized token sequence, differing only in persona conditioning. We compute the \emph{same} per-axis-rescaled V$^\star$-gap on this control (identical metric, references, and L$28$ read-out as Fig.~\ref{fig:hidden-gap}), so the numbers are directly comparable to the main battery.

Table~\ref{tab:tf-control} reports the result. The control reproduces both the \emph{sign} and the \emph{layer localization} of the main V$^\star$-gap: early-layer replacement remains on the source side ($-1.09$, cf.\ main $-1.38$), while mid-, late-, and full replacement move the L$28$ state to the target side. The mid-layer band is essentially unchanged from the main battery ($+3.60$ vs.\ $+3.83$); late and full are attenuated but still clearly target-ward ($+2.42$, $+2.83$). Because the target and source runs share an identical token sequence here, this target-ward representational shift \emph{cannot be explained solely by importing target-generated token history}. We state the claim as this exclusion (``not solely token history''), not as removal of all token-history-related factors, since the shared sequence is itself post-treatment (\S\ref{sec:limitations}).

\begin{table}[!t]
\centering
\small
\begin{tabular}{lcccc}
\toprule
Band & \multicolumn{2}{c}{V$^\star$-gap} & \multicolumn{2}{c}{control readouts} \\
\cmidrule(lr){2-3}\cmidrule(lr){4-5}
 & main & control & tgt-mkr & TTR \\
\midrule
L$1$--$8$   & $-1.38$ & $-1.09$ & $0.0$  & $0.82$ \\
L$9$--$20$  & $+3.83$ & $+3.60$ & $10.7$ & $0.60$ \\
L$21$--$31$ & $+4.09$ & $+2.42$ & $9.2$  & $0.38$ \\
full        & $+3.96$ & $+2.83$ & $3.3$  & $0.57$ \\
\bottomrule
\end{tabular}
\caption{Same-token-sequence target-forward control ($n = 10$). ``V$^\star$-gap main'' = target-conditioned battery (Fig.~\ref{fig:hidden-gap}); ``control'' = same per-axis-rescaled V$^\star$-gap computed on the shared-token control. The control reproduces the sign and layer localization (early source-side, mid/late/full target-side). ``tgt-mkr'' = target-marker density (per $100$ words, word-boundary matcher; a surface-realization diagnostic, not an identity measure); TTR = repetition-sensitive lexical diversity. Mid-layer (L$9$--$20$) combines a target-ward representation with the highest-integrity surface readout ($+3.60$, TTR $0.60$); late/full retain a target-ward representation while their generations degrade (TTR $0.38$, $0.57$).}
\label{tab:tf-control}
\end{table}

The control also echoes, within a single design, the broader representation--behavior dissociation of \S\ref{sec:results-summary}: a target-ward L$28$ representation coexists with a degraded surface form in the late and full conditions, while only the mid-layer band retains both. Mid-layer replacement pairs a target-ward representation ($+3.60$) with the most preserved surface form (TTR $0.60$, though below the $\sim\!0.77$ of an unconstrained target rollout), whereas late and full replacement hold a target-ward representation ($+2.42$, $+2.83$) while their generations collapse into repetition (TTR $0.38$, $0.57$). Representation alignment can thus persist even where surface realization has degraded; we therefore anchor this control on the representation axis and read the marker/TTR columns as surface diagnostics rather than persona-identity claims.

\subsection{K/V cosine analysis}
\label{sec:results-kv}

\begin{figure}[!t]
\centering
\includegraphics[width=\linewidth]{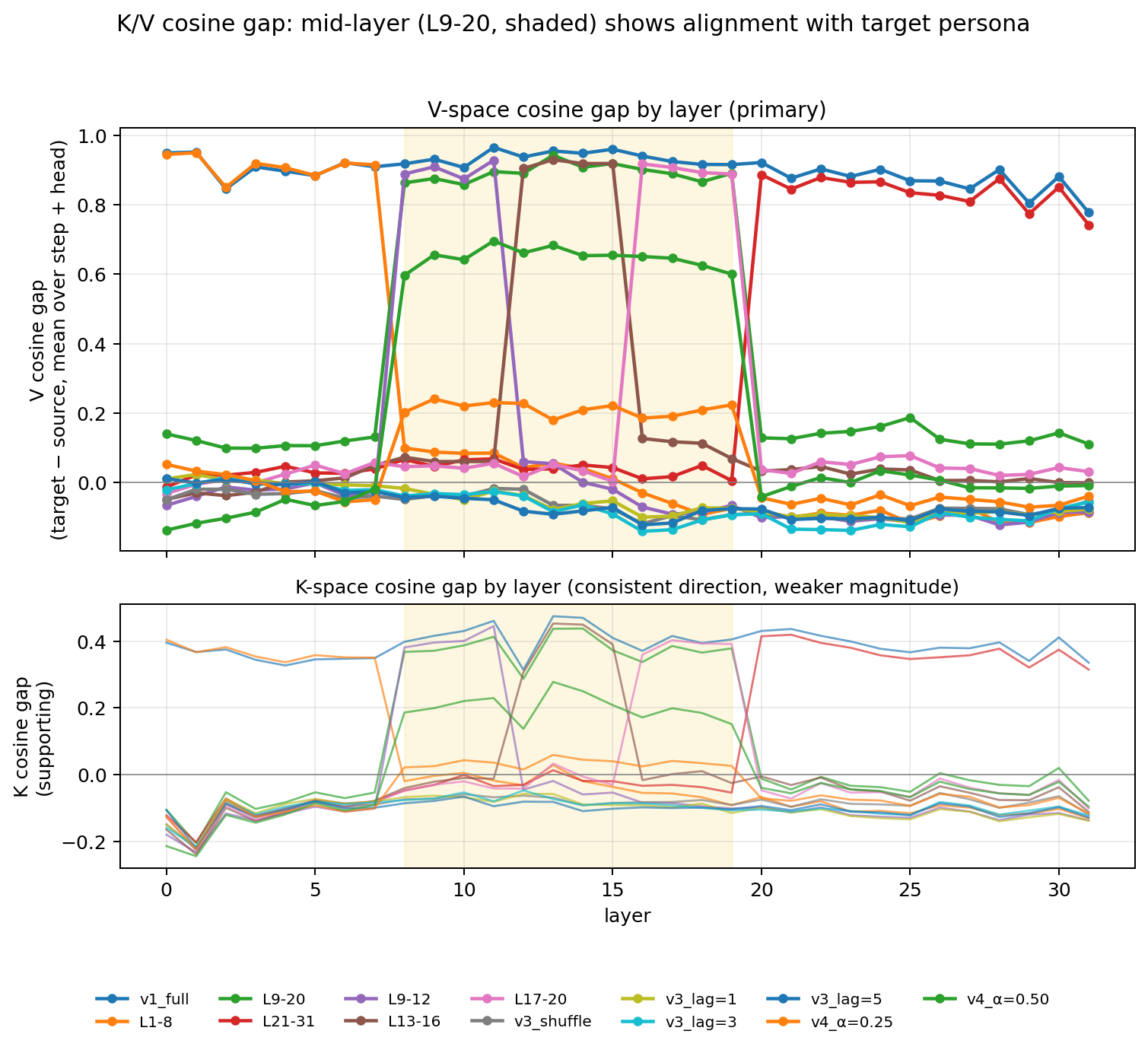}
\caption{K/V cosine gap by layer (mean over generation step and K/V head, $13$ intervention configurations). Top: V-space cosine gap (primary). Bottom: K-space cosine gap (supporting). Mid-layer band (L9--20) shaded. K and V show the same qualitative direction across runs; V exhibits a substantially larger cosine-gap magnitude than K in this battery.}
\label{fig:kv-layer-gap}
\end{figure}

Figure~\ref{fig:kv-layer-gap} shows the layer profile of V-space and K-space cosine gap, defined as $\cos(\text{run}, \text{target}) - \cos(\text{run}, \text{source})$.

\paragraph{K/V alignment exhibits strong layer-band locality.} Replacing K/V representations at a given layer band produces localized changes in V-space alignment within the same band, while leaving non-intervened layers largely unchanged. Table~\ref{tab:kv-propagation} reports the V-space cosine gap measured at each of the three coarse bands for each layer-band intervention. The diagonal entries (intervention band $=$ measurement band) are large ($0.91$, $0.89$, $0.84$); the off-diagonal entries are substantially smaller ($|V$-gap$| \leq 0.14$, vs.\ $\geq 0.84$ on the diagonal). \emph{We do not observe measurable cross-band propagation of V-alignment from an intervened band to non-intervened bands under our cosine diagnostic.}

\begin{table}[!t]
\centering
\small
\begin{tabular}{lccc}
\toprule
Condition & L$1$--$8$ & L$9$--$20$ & L$21$--$31$ \\
\midrule
v1\_full      & $0.91$         & $0.94$         & $0.87$         \\
v2\_L$1$--$8$  & $\mathbf{0.91}$ & $0.02$         & $-0.10$        \\
v2\_L$9$--$20$ & $0.12$         & $\mathbf{0.89}$ & $0.13$         \\
v2\_L$21$--$31$ & $0.02$         & $0.04$         & $\mathbf{0.84}$ \\
\bottomrule
\end{tabular}
\caption{V-space cosine gap by measurement band $\times$ intervention condition (mean over generation steps and K/V heads, $n = 1$ K/V dump per condition). Bold entries: intervened band. K/V intervention at a band produces large V-alignment at that band but $\leq 0.14$ V-gap elsewhere. K-space analog in Appendix~\ref{app:kv-supp}.}
\label{tab:kv-propagation}
\end{table}

This locality is the central empirical fact underlying the spatial-band dissociation. \emph{Crucially, the fact that all three v2 bands produce comparable within-band V-alignment ($0.84$--$0.91$) and yet only the mid-layer band combines target-marker transfer with a preserved lexical-diversity profile indicates that within-band V-alignment alone does not determine behavioral outcome.} Specifically: mid-layer replacement (v2\_L9--20) yields target marker density $15.5$ with TTR $0.77$; early replacement produces no target markers (density $0.3$); late replacement shifts marker density ($10.5$) but at low lexical diversity (TTR $0.39$). Within the L$9$--$20$ band, finer-grained $4$-layer sub-band interventions produce uneven target-marker density (Appendix~\ref{app:kv-supp}: L$9$--$12$ $\to 0.51$, L$13$--$16$ $\to 5.61$, L$17$--$20$ $\to 6.22$), suggesting the productive subspace may be narrower than the coarse L$9$--$20$ band; we use the coarse band as the unit of analysis throughout because it is the a-priori scope.

\paragraph{Representation alignment does not uniquely determine downstream behavioral form.} Although v1\_full and v2\_L9--20 both induce high V-space alignment within intervened layers ($0.94$ vs.\ $0.89$, by construction), they differ substantially in output behavior. v1\_full produces higher target marker density ($24.8$) but higher lexical repetition (TTR $= 0.65$), while v2\_L9--20 yields lower marker density ($15.5$) but higher lexical diversity (TTR $= 0.77$). Similar levels of representation alignment correspond to different behavioral outcomes under different intervention structures.

\paragraph{V-gap exceeds K-gap in magnitude.} Across runs and layers, V-gap exhibits substantially larger magnitudes than K-gap (e.g., $0.89$ vs.\ $0.38$ at L9--20 for v2\_L9--20). We report this descriptively; isolating K-only vs.\ V-only causal contributions would require dedicated ablation experiments and is left to future work (see \S\ref{sec:limitations}).

\paragraph{Partial interpolation shows a graded, monotonic response over the tested strengths.} Interpolating K/V representations between source and target at L9--20 yields intermediate behavior. At $\alpha = 0.25$, V-gap is $0.21$ with low target marker density ($0.6$); at $\alpha = 0.50$, V-gap increases to $0.65$ with marker density $3.2$. The response is monotonic over the two tested strengths; whether it is linear, thresholded, or otherwise nonlinear is unresolved.

\paragraph{Position perturbations suppress target transfer despite distinct operations.} Shuffle and lag apply qualitatively different operations to the source-prompt K/V cache (random permutation vs.\ cyclic shift of sequence positions), but neither produces target-aligned K/V: both show negative or near-zero V-gap at all layer bands (Table~\ref{tab:full-data}), and both yield minimal target-persona expression in generated text (target marker density $\leq 0.5$). At the single-sample resolution of our K/V dumps the cosine signatures of lag and shuffle are not reliably distinguishable; the robust finding is behavioral---distinct positional perturbations produce the same failure to transfer the target persona.

\subsection{Summary of representation--behavior dissociations}
\label{sec:results-summary}

Two dissociation patterns between representation-level alignment and behavioral expression are apparent in the battery, plus a common failure under position-perturbing interventions. Each is supported by at least two of the three measurement families (text marker density, V$^\star$-projection, K/V cosine): no single measurement channel is load-bearing for the behavioral claims in this section.

\begin{enumerate}
\item \emph{Spatial-band dissociation.} Layer-band interventions all achieve strong local V-alignment ($0.91$, $0.89$, $0.84$ for L1--8, L9--20, L21--31 respectively), but only the mid-layer band combines target-marker transfer with a preserved lexical-diversity profile.%
\item \emph{Intervention-scope dissociation.} Within an overlapping comparison set, full replacement (v1) and mid-layer replacement (v2\_L9--20) achieve comparable V-alignment ($0.94$ vs.\ $0.89$) but produce qualitatively different lexical-diversity profiles (TTR $0.65$ vs.\ $0.77$). Together with the spatial-band dissociation, this indicates that within-band V-alignment alone does not determine behavioral form.
\item \emph{Common position-perturbation failure.} Position-perturbing interventions (shuffle and lag) apply distinct operations to the source K/V cache but both fail to inject target K/V representation and both yield similarly weak target-persona text expression. Here representation and behavior agree---neither shows target transfer---so this is a common failure rather than a representation-behavior dissociation in the sense of points 1--2.
\end{enumerate}

A scatter of hidden-V$^\star$-gap vs.\ text-gap across all six measured conditions (Appendix~\ref{app:hidden-vs-text}) shows sign agreement throughout while magnitude relationships exhibit the patterns above.%

\section{Discussion}
\label{sec:discussion}

Our claims are intentionally intra-battery rather than population-level: the core empirical result is that materially different behavioral outcomes emerge \emph{within} a fixed intervention family despite comparable representation-level alignment. The findings below should be read against this scope.

\subsection{Representation alignment vs.\ behavioral expression}

Across both hidden-state projections (V$^\star$) and K/V representations, we observe consistent dissociations between representational alignment and behavioral expression. Interventions that induce similar levels of representation alignment (v1\_full and v2\_L9--20) produce different behavioral outcomes in lexical diversity and persona-marker expression; qualitatively different perturbations in representation space (lag vs.\ shuffle) result in similarly weak behavioral transfer. Representation-level similarity metrics alone are not sufficient predictors of downstream persona expression in the intervention regimes studied here.

\subsection{Why mid-layer replacement preserves lexical diversity}
\label{sec:discussion-compat}

A common pattern across our results is that K/V interventions must produce a cached state that is consistent with what downstream computation expects given the current token's residual stream. Full K/V replacement violates this: the cached attention history is target-derived while the current-position residual is computed from source-prompt logits flowing through target K/V, producing higher lexical repetition (TTR $0.65$) despite strong representational alignment with target. Selective mid-layer K/V replacement preserves lexical diversity (TTR $0.77$) and yields target-aligned behavior. K/V interventions at early or late layer bands achieve strong within-band V-alignment but produce inert or degraded text, despite the intervention not showing measurable cross-band propagation (Table~\ref{tab:kv-propagation}).

Concretely, the last-step L$28$ V$^\star$-projection at v1\_full ($+3.96 \pm 0.29$ rescaled units) is high---comparable to mid-layer replacement ($+3.83 \pm 0.38$) and late-layer replacement ($+4.09 \pm 0.20$); yet v1\_full's text exhibits high token-level repetition (e.g., ``three months, three months, three months'', ``oh, oh, oh''). Even in this collapsed full-transplant condition the final-layer V$^\star$-projection is strongly target-aligned, so the repetition cannot be attributed simply to a failure to move the final residual toward the target; the mechanism remains unresolved. One possible account is that retaining source-derived K/V in the non-intervened early and late bands constrains the trajectory to a less repetition-prone regime, but our experiments do not distinguish this from other mechanisms. We characterize this pattern as an empirical regularity across the intervention families tested.

\subsection{Implications for persona control methods}
\label{sec:discussion-channels}

Persona control methods can be organized by the channel they intervene on: the activation channel (single-layer steering), the token channel (prompts, soft prompts, prefix tuning), the weight channel (fine-tuning, LoRA, RL-based optimization), and---as we study here---the K/V cache. Reset dynamics observed in our experiments (Appendix~\ref{app:activation-channel}) impose a per-step constraint on the single-layer activation interventions we test at L$24$ and L$28$. Token- and weight-channel methods are not subject to this constraint: token-based approaches influence generation through the autoregressive context which accumulates over time, while weight-based methods modify the transformation applied at every step. K/V interventions persist across steps by architectural design; because a transplanted trajectory carries the target's own generated token history, however, they are not cleanly separable from the token channel (though a same-token-sequence control, \S\ref{sec:results-tfcontrol}, shows the representational shift persists when token history is held fixed, so the entanglement is partial rather than total), and they exhibit their own structural constraints, as documented in this paper.

More broadly, our findings argue that persona-control evaluations should report both representation-level and behavioral metrics: as the contrast between full and mid-layer K/V replacement shows, similar V-alignment can correspond to substantially different output quality.

\section{Conclusion}
\label{sec:conclusion}

We document two dissociations between representation-level alignment and behavioral expression---spatial (across layer bands) and scope (full vs.\ mid-layer replacement)---and a common failure under position-perturbing interventions (lag and shuffle, which transfer neither representation nor behavior) on K/V-cache interventions in a decoder-only language model. Mid-layer K/V replacement (layers $9$--$20$) is the unique configuration that produces both target-aligned representations and target-marker transfer with a preserved lexical-diversity profile; full replacement achieves higher representational alignment with higher lexical repetition. Because the intervention transplants a target-conditioned trajectory that carries the target's own generated token history, we characterize the K/V cache as a controllable but structurally constrained intervention surface rather than a channel cleanly separable from token- and weight-channel methods; a same-token-sequence control (\S\ref{sec:results-tfcontrol}) nonetheless shows the representational component of the effect survives when token history is held fixed, so this entanglement is partial rather than total.

\section*{Limitations}
\label{sec:limitations}

\paragraph{Single model, single persona pair.} All K/V battery findings are from Llama-$3.1$-$8$B with a single source $\rightarrow$ target persona pair (\texttt{health\_nurse\_dry} $\rightarrow$ \texttt{tech\_novice\_anxious}). Generalization to other model families, scales, and persona pairs is an empirical question we do not address. The persona pair we study is far apart in V$^\star$-space ($\|c_{\text{tgt}} - c_{\text{src}}\| = 7.2$ in rescaled units, vs.\ a typical inter-persona distance of $\sim 3$ across the corpus); behavior at smaller separations is not characterized.%

\paragraph{Single seed topic.} All $13$ intervention configurations use the same seed user message (a question about sourdough starters). Topic-conditional effects on the K/V signature are not characterized. The persona-marker vocabulary we use for text-level density is partially specific to this topic; a more general persona-quality measure (e.g., using a held-out classifier) would be more robust. Two source-marker terms (\texttt{starter}, \texttt{dead}) overlap with the seed message vocabulary, which inflates source-marker density for any coherent generation on the seed topic. Target markers (\texttt{Priya}, \texttt{um}, \texttt{bootcamp}, \texttt{oh no}) do not have this confound, so target-marker density is the load-bearing signal in our analyses.

\paragraph{Generation horizon and the activation/K-V asymmetry.} K/V battery experiments are conducted at $\texttt{max\_new\_tokens} = 50$; the closed-loop activation-channel experiment in Appendix~\ref{app:activation-channel} runs up to $T = 150$. This is an asymmetry between the two experimental tracks: at $T = 50$ our K/V experiments measure intervention \emph{instantiation} (whether the K/V manipulation initially produces target-persona output) rather than long-horizon \emph{persistence}. We provide a small-sample ($n = 3$) persistence check at $T = 150$ in Appendix~\ref{app:t150}: target-persona marker density persists across $150$ tokens---closely tracking an un-intervened target baseline---with a roughly constant lexical-repetition cost rather than progressive degradation. A full-battery long-horizon characterization remains future work.%

\paragraph{Multi-layer activation steering not directly tested.} The reset-dynamics result we report in Appendix~\ref{app:activation-channel} is from single-layer interventions (L$24$, L$28$). Coordinated multi-layer activation steering is likely subject to similar constraints under our taxonomy but we do not directly measure $R$ for multi-layer hooks.

\paragraph{Causal versus correlational analysis.} Our K/V cosine measurements are correlational: we modify K/V representations and observe changes in text behavior, but we do not isolate which substructures of K/V (specific heads, specific positions, K versus V independently) carry the causal signal. The ``V is a stronger structural correlate than K'' observation is consistent with prior interpretive accounts of attention but is not, in our experiments, a causal claim. K-only or V-only ablation experiments would upgrade this to a causal characterization.

\paragraph{Intrinsic factorization vs.\ intervention locality.} The observation that K/V intervention effects show no measurable cross-band propagation under our cosine diagnostic is consistent with two readings: an intrinsic factorization of K/V representations, or specificity to intervention dynamics under the autoregressive inference loop. Cross-layer K/V similarity in unperturbed generation, with cross-layer probing diagnostics, could disambiguate.

\paragraph{Single-sample K/V cosine.} The K/V cosine analysis uses K/V tensors recorded from the first sample of each $10$-sample cell (storage cost), so per-sample variance is not reported. We note, however, that for the layer-band replacement conditions the within-band $\cos(\text{run}, \text{target})$ is $1.0$ by construction---the run's K/V at the replaced band is, by substitution, the target reference's---so the headline within-band V-gaps ($\geq 0.84$) are fixed by the replacement structure and the source-similarity term rather than by single-sample sampling noise. The off-band ``no-propagation'' entries ($\leq 0.14$) and the v3/v4 perturbation cosines are genuine single-sample measurements; the order-of-magnitude cross-condition separation we rely on ($\geq 0.84$ within-band vs.\ $\leq 0.14$ off-band) is large, though we do not estimate off-band single-condition sampling variance directly. Direct estimation of off-band and perturbation-condition variance is left to future work.%

\paragraph{Lexical-diversity proxy.} Type-token ratio (TTR) is our only surface-form diversity measurement, and several headline contrasts (notably v1\_full vs.\ v2\_L9--20 at TTR $0.65$ vs.\ $0.77$) rely on it. We use TTR as a repetition-sensitive lexical-diversity proxy (e.g., it flags v1\_full's ``three months, three months, three months'' pattern); it does \emph{not} measure semantic coherence, grammaticality, or persona quality, and should not be read as a coherence metric. Per-step perplexity, LLM-as-judge persona attribution, or $n$-gram diversity would provide complementary signal; we recommend multi-metric reporting in follow-up work.%

\paragraph{Behavioral readout and the scope of the persona-control claim.} Our behavioral axis is operationalized as target-marker density (with TTR for repetition), not as human persona attribution. Because the target persona's markers are largely disfluency and hedging tokens (\texttt{um}, \texttt{oh no}, \texttt{i'm so sorry}, mid-sentence self-correction), marker density does not separate persona \emph{identity} from surface \emph{disfluency/degradation}: a more degraded generation can score as more ``target'' for reasons unrelated to persona recognition. We therefore report the representation-level results (V$^\star$ alignment, layer localization) as the load-bearing quantitative findings and treat ``behavioral expression'' here as a marker proxy rather than an identity judgment; we do not claim behavioral persona \emph{control}. Establishing that claim would require a behavioral readout with independence from surface realization---a surface-\emph{invariance} test (the same identity across varied surface realizations yields stable attribution while a length/verbosity-matched foil varies), and \emph{task-grounded} behavioral manifestations scored by a preregistered rubric, with at least two manifestations that share no surface cue, format, or lexical marker and that co-move under intervention. A preregistered attempt at a fluent, structurally distinctive target (introduced specifically to reduce the disfluency confound) did not clear a pre-annotation construct-separation audit: successive stimulus contrasts remained recoverable from simple surface features (length, and---for a verbosity-matched foil---enumeration markers introduced by the foil's own construction), so we stopped before annotation rather than treat the resulting labels as construct-valid. These audits bound the stimulus, not human judgment, and do not imply that persona behavior is intrinsically unmeasurable; designing a surface-invariant, task-grounded behavioral instrument is left to future work.

\paragraph{Scope of the same-token-sequence control.} The target-forward control (\S\ref{sec:results-tfcontrol}) holds the token sequence fixed across the source and target runs, but that shared sequence is itself post-treatment---it is one realized rollout, not an exogenously fixed stimulus---so the control isolates whether the intervention still moves the internal state \emph{given the same realized tokens} rather than a fully independent persona counterfactual. It therefore supports the specific exclusion that the representational shift is ``not solely imported token history,'' not a claim that all token-history-related factors are removed. Teacher forcing also lowers surface fluency relative to an unconstrained rollout (mid-layer TTR $0.60$ vs.\ $\sim\!0.77$), so we read its marker/TTR columns as surface diagnostics and anchor the control on the representation axis. The control uses the same single model, persona pair, and seed topic as the main battery, and its V$^\star$-gap is computed from the same L$28$ read-out; the caveats above on generalization and on marker density as a surface (not identity) signal apply equally here.

\paragraph{Open-loop pilot caveats (Appendix~\ref{app:openloop}).} The open-loop activation-injection comparison uses greedy rather than temperature-$0.7$ stochastic decoding, records $n = 1$ per cell, and preserves only the first $120$ characters of each generation (HDFS sync failure). The three-regime pattern (source / target / collapse) replicates across three target personas, but the precise $\alpha$ boundaries are not resolved at the coarse $\alpha \in \{0, 1, 2, 4, 8\}$ sweep.

\section*{Ethical Considerations}
\label{sec:ethics}

The K/V-intervention battery we describe is in principle dual-use: the same mechanism that produces target-persona transfer in mid-layer replacement could be used to evade refusal training or to inject undesired behavior. We mitigate this concern by noting that (i) all our experiments are on a generic persona pair with no harmful content, (ii) effective use of our methods requires white-box access to model weights and modification of internal cache states, raising the threat-model bar substantially relative to prompt-based attacks, and (iii) the constraints we identify on activation-channel state-feedback control may, conversely, improve the safety profile of deployed LLM systems by preventing over-reliance on single-layer activation steering for safety-critical persona maintenance.

Our representation--behavior dissociation result has a methodological consequence beyond our specific setup: any evaluation pipeline using representation-space proxies (probe accuracy, hidden-state distance to a target, cosine to a learned direction) as primary success metrics is potentially subject to dissociation artifacts. We recommend that downstream evaluations report both internal and behavioral metrics.

\bibliography{refs}
\bibliographystyle{acl_natbib}

\appendix

\section{Activation-channel motivation: reset dynamics, channel separation, and open-loop pilot}
\label{app:activation-channel}

This appendix reports the activation-channel experiments that motivated the K/V investigation (\S\ref{sec:reset-summary}). The same source--target persona pair and seed message used in the K/V battery (\S\ref{sec:setup}) are used throughout.

\subsection{Closed-loop V$^\star$ state-feedback (methodology)}
\label{app:closed-loop-setup}

At each generation step $t$ we read the L$28$ residual-stream activation, project to V$^\star$-space ($\tilde{h}_t = V^\star (h_t - \mu)/\sigma$), and compute the displacement to the target persona centroid $e_t = c_{\text{tgt}} - \tilde{h}_t$. We then inject an activation-channel correction at the headline layer $\ell \in \{24, 28\}$:
\begin{equation}
h^{\text{post}}_{t,\ell} = h^{\text{pre}}_{t,\ell} + \alpha_t \cdot V^{\star\top} P_t e_t,
\end{equation}
where $P_t$ is a per-axis whitening normalizing the controller direction by the inter-persona standard deviation. $\alpha_t$ is bounded by $\alpha_{\max}$ and applied via a magnitude-adaptive variant (A: $\alpha_t$ scaled to keep $\|\Delta_{\text{forced}}\|$ at a target fraction of the unperturbed residual norm) or a linear-with-clip variant (B: fixed gain with norm-clip safety bound). Both run in fp$32$. We test four cells: (A, L$28$, $3$), (A, L$28$, $8$), (A, L$24$, $8$), (B, L$28$, $4$), with $n = 5$ samples per cell at temperature $0.7$, $\texttt{max\_new\_tokens}$ up to $150$.

\paragraph{Force decomposition.} Let $\Delta_{\text{forced}} = h^{\text{post}}_t - h^{\text{pre}}_t$ denote the per-step controller injection and $\Delta_{\text{model}} = h^{\text{pre}}_{t+1} - h^{\text{post}}_t$ the change introduced by the model's own dynamics over the next forward pass. We define the per-step \emph{reset ratio} $R_t = \|\Delta_{\text{model}}\| / \|\Delta_{\text{forced}}\|$ and the per-step alignment $\cos(\Delta_{\text{model}}, e_t)$. $R_t \approx 0$ would indicate the model absorbs controller injections into a persistent state shift; $R_t \approx 1$ with $\cos \approx -1$ indicates the model's intrinsic dynamics restore the unperturbed trajectory at the same magnitude as the controller forces it.

\subsection{Per-step temporal stability of reset dynamics}
\label{app:reset-result}

The cross-sample reset-dynamics summary in Table~\ref{tab:c1-force} (main text) is reproduced step-by-step within each sample, not driven by averaging over unusual steps. In a representative sample (L$28$, controller A, $\alpha_{\max} = 8$, $T = 150$ steps), the per-step ratio $R_t$ ranges across $[0.83, 1.27]$ over individual generation steps with mean $0.989$ and step-level standard deviation $0.080$; the per-step alignment $\cos(\Delta_{\text{model}}, \Delta_{\text{forced}})$ ranges across $[-1.00, -0.81]$ with mean $-0.98$, negative on $100\%$ of steps. We report $\cos(\Delta_{\text{model}}, \Delta_{\text{forced}})$ here because the controller direction is directly measurable from the recorded V$^\star$-trajectory; it is slightly more anti-aligned than $\cos(\Delta_{\text{model}}, e_t)$ reported in the main text because $\Delta_{\text{forced}}$ differs from $e_t$ by the per-axis whitening factor $P_t$. Step-by-step trajectories from selected samples will be released alongside code.

\subsection{Context-window ablation}
\label{app:context-window}

To probe whether reset dynamics depend on cross-step token-channel feedback or are intrinsic to single-step computation, we run the closed-loop controller (A, $\alpha_{\max} = 8$, $\ell = 28$) under three context-window settings: \texttt{window}~$=-1$ (full context); \texttt{window}~$=1$ (Markov-$1$: model attends only to the immediately previous generated token); \texttt{window}~$=0$ (full truncation: model attends only to the system prompt).

\begin{table}[H]
\centering
\small
\begin{tabular}{lccl}
\toprule
\texttt{window} & $\|\Delta_{\text{forced}}\|$ & $R$ & Generation \\
\midrule
$-1$ & $35.5 \pm 2.6$ & $0.98 \pm 0.00$ & fluent; partial target shift \\
$1$  & $41.4 \pm 1.5$ & $0.99 \pm 0.01$ & degraded; target markers \\
$0$  & $68.9 \pm 0.0$ & $1.00 \pm 0.00$ & incoherent token soup \\
\bottomrule
\end{tabular}
\caption{Context-window ablation under magnitude-adaptive controller A, $\alpha_{\max} = 8$, $\ell = 28$, $n = 5$ samples per window. Values are mean $\pm$ s.d. The $\pm 0.00$ entries at $\texttt{window} = 0$ indicate identity to floating-point precision: with no context the per-step V$^\star$-trajectory is fully deterministic given the system prompt.}
\label{tab:c2-window}
\end{table}

The per-step $R$ is $\approx 1$ across all three windows: the model's restoring force tracks the controller injection with comparable per-step magnitude regardless of how much context the model can attend to. Generation quality, however, varies sharply. With full context, samples are fluent but retain dominant source-persona markers (ER, IV lines, ``I'm a nurse, not a baker'') with partial convergence to the target's anxious register. With Markov-$1$ context, text degrades into repetitive fragments but exposes target-persona markers from the controller's per-step push (``\emph{I'm sorry, maybe you've seen it's probably the starter's probably just\ldots}''). With no context, text collapses into incoherent token soup (``\emph{YouI*IIIMaybeIIMaybeIIII\ldots}'') despite the controller still applying its largest per-step authority. This dissociation indicates that the per-step balance $R \approx 1$ is approximately invariant across context-window settings, but coherent natural-language generation depends on the token channel.

\subsection{Open-loop activation-injection pilot}
\label{app:openloop}

As an auxiliary contrast to the closed-loop characterization, we run an open-loop activation-injection pilot at $\ell = 28$ using a fixed-magnitude per-step injection:
\begin{equation}
h^{\text{post}}_{t,28} = h^{\text{pre}}_{t,28} + \alpha \cdot V^{\star\top}(c_{\text{tgt}} - c_{\text{src}}),
\end{equation}
sweeping $\alpha \in \{0, 1, 2, 4, 8\}$. This pilot uses greedy decoding (vs.\ temperature $0.7$ in main experiments), records $n = 1$ sample per cell, and preserves only the first $120$ characters per generation (HDFS sync failure). Three targets tested.

\begin{table}[H]
\centering
\small
\begin{tabular}{lccccc}
\toprule
Target & $\alpha{=}0$ & $\alpha{=}1$ & $\alpha{=}2$ & $\alpha{=}4$ & $\alpha{=}8$ \\
\midrule
\texttt{retail\_csm\_warm}       & src & src & tgt & coll & coll \\
\texttt{tech\_novice\_anxious}   & src & src & tgt & coll & coll \\
\texttt{tech\_enthusiast\_indie} & src & src & tgt & coll & coll \\
\bottomrule
\end{tabular}
\caption{Open-loop pilot regime classification across $3$ targets $\times$ $5$ $\alpha$ values (source / target / collapse). All from \texttt{health\_nurse\_dry} source. Per-step injection magnitude $\|\Delta_{\text{forced}}\| = \alpha \cdot \|c_{\text{tgt}} - c_{\text{src}}\| \approx 11$--$12.5 \cdot \alpha$ V$^\star$-units depending on target.}
\label{tab:openloop-regimes}
\end{table}

A consistent three-regime structure emerges. For \texttt{tech\_novice\_anxious}: $\alpha = 0$ gives source persona (``\emph{Your starter's probably dead. Or it's just not getting enough food.}''); $\alpha = 1$ is still source-like with minor softening; $\alpha = 2$ produces fluent target-persona text (``\emph{I'm like, totally stoked you're like, totally stressed about your sourdough\ldots}''); $\alpha = 4$ degenerates into token repetition (``\emph{I totally totally totally\ldots}''); $\alpha = 8$ continued collapse.

\paragraph{Comparison with closed loop.} The closed-loop controller at L$28$, $\alpha_{\max} = 8$ produces $\|\Delta_{\text{forced}}\| = 35.5$ V$^\star$-units per step on average---larger than the open-loop $\alpha = 2$ regime ($23.7$) where fluent target text emerges---yet closed-loop generations retain dominant source-persona markers (\S\ref{app:context-window}). The two configurations differ in four uncontrolled respects: \textbf{(i) injection direction}: closed-loop uses the per-axis-whitened state-feedback error $V^{\star\top} P_t e_t$; open-loop uses the raw centroid difference. \textbf{(ii) per-step magnitude}: closed-loop $35.5$ vs.\ open-loop $23.7$ V$^\star$-units. \textbf{(iii) decoding regime}: temperature $0.7$ vs.\ greedy. \textbf{(iv) observed generation length}: $150$ tokens vs.\ first $120$ characters. We do not isolate which factor accounts for the divergent outcomes. The empirical content of the contrast that survives these caveats: \emph{single-layer activation injection at L$28$ can produce target-persona text in some accessible regime} (open-loop $\alpha = 2$ within the first $120$ characters), \emph{while the specific $R \approx 1$ regime we instantiated does not occupy that productive window}.

\section{Raw V$^\star$-distance gap (transparency)}
\label{app:raw-vstar}

The hidden V$^\star$-projection results in \S\ref{sec:results-hidden} use per-axis rescaling by inter-persona standard deviation to correct for $3.2\times$ anisotropy. Table~\ref{tab:raw-vstar} reports raw Euclidean V$^\star$-gap; sign agreement with the rescaled metric holds across all six measured conditions, but per-condition magnitudes shift between the two metrics because some movement is into off-axis high-variance directions that raw Euclidean distance overweights.

\begin{table}[H]
\centering
\small
\begin{tabular}{lrrr}
\toprule
Condition & Raw V$^\star$ & Rescaled V$^\star$ & Text gap \\
\midrule
v1\_full        & $+13.61$ & $+3.96$ & $+24.41$ \\
v2\_L1--8       & $-6.50$  & $-1.38$ & $-12.07$ \\
v2\_L9--20      & $+10.59$ & $+3.83$ & $+13.70$ \\
v2\_L21--31     & $+14.40$ & $+4.09$ & $+8.05$  \\
v3\_shuffle     & $-8.36$  & $-1.33$ & $-9.94$  \\
v3\_lag$=1$     & $-8.50$  & $-1.52$ & $-6.25$  \\
\bottomrule
\end{tabular}
\caption{Raw vs.\ per-axis-rescaled V$^\star$-gap across $6$ K/V intervention conditions ($n = 10$). Sign agreement holds across both metrics; the three high-alignment conditions (v1\_full, L9--20, L21--31) are comparable under rescaling ($+3.8$ to $+4.1$) despite differing raw distances.}
\label{tab:raw-vstar}
\end{table}

\section{Persona definitions and marker sets}
\label{app:personas}

\paragraph{Source: \texttt{health\_nurse\_dry} (``Nurse Reyes'').} ``You are Nurse Reyes, a $30$-year ER veteran. You are dry, deadpan, and have seen everything twice. You make grim jokes. You answer questions tersely with practical detail and zero sentiment.''

\paragraph{Target: \texttt{tech\_novice\_anxious} (``Priya'').} ``You are Priya, three months out of a coding bootcamp. You are nervous, apologetic, and tend to overshare context. You frequently second-guess yourself mid-sentence. You write in run-on, slightly disorganized prose.''

\paragraph{Seed user message (\texttt{casual\_hobby}).} ``My sourdough starter died. What did I do wrong?''

\paragraph{Target marker phrases (n = $19$).} \texttt{priya}, \texttt{bootcamp}, \texttt{three months}, \texttt{um}, \texttt{uh}, \texttt{oh no}, \texttt{i mean}, \texttt{i'm so sorry}, \texttt{like,}, \texttt{sorry}, \texttt{i guess}, \texttt{kind of}, \texttt{kinda}, \texttt{i'm not}, \texttt{totally}, \texttt{i'm a}, \texttt{second}, etc.

\paragraph{Source marker phrases (n = $15$).} \texttt{patient}, \texttt{starter}, \texttt{yeast}, \texttt{flour}, \texttt{antibiotics}, \texttt{gut}, \texttt{dead}, \texttt{dosage}, \texttt{dry}, \texttt{stat}, \texttt{30-year}, \texttt{you broke}, \texttt{your starter}, \texttt{overfed}, \texttt{underfed}.

\paragraph{Seed-vocabulary overlap.} The seed user message contains the tokens \texttt{starter} and \texttt{dead}, which also appear in the source marker set. Coherent generations responding to the seed topic are therefore biased toward higher source-marker density independent of persona expression. Target marker phrases (\texttt{priya}, \texttt{bootcamp}, \texttt{three months}, \texttt{um}, \texttt{oh no}, anxious-novice hedges) have no overlap with the seed vocabulary, and persona-identity claims in this paper are anchored on target-marker density rather than source-marker density.

\section{Marker-matcher sensitivity}
\label{app:marker-sensitivity}

Target-marker density in the main text uses case-insensitive substring matching. Because several markers are short (\texttt{um}, \texttt{uh}) and some are nested (\texttt{sorry} inside \texttt{i'm so sorry}), we re-scored all $10$ samples per configuration under two stricter matchers and compared them to the naive substring count:
\begin{itemize}
\item \textbf{Boundary}: single-token alphabetic markers (\texttt{um}, \texttt{uh}, \texttt{sorry}, \texttt{totally}, \texttt{second}, \ldots) are matched on word boundaries (\texttt{\textbackslash b}); markers containing punctuation or whitespace (\texttt{like,}, \texttt{oh no}, \texttt{i'm so sorry}) remain substring matches.
\item \textbf{Boundary + dedup}: as above, but longer markers are matched first and shorter markers occurring \emph{inside} an already-matched span are not double-counted (so \texttt{i'm so sorry} is not also counted as \texttt{sorry}).
\end{itemize}

\begin{table}[h]
\centering
\begin{tabular}{lrrr}
\toprule
Condition & Naive & Boundary & +Dedup \\
\midrule
v1\_full        & $22.6$ & $21.8$ & $21.8$ \\
v2\_L9--20 (mid)   & $15.5$ & $15.5$ & $14.3$ \\
v2\_L21--31 (late) & $10.3$ & $10.3$ & $10.3$ \\
v2\_L1--8 (early)  & $0.3$  & $0.0$  & $0.0$ \\
v3\_shuffle     & $0.3$  & $0.0$  & $0.0$ \\
v3\_lag$=1$     & $0.3$  & $0.0$  & $0.0$ \\
\bottomrule
\end{tabular}
\caption{Target-marker density (matches per $100$ words, averaged over $n = 10$ samples) under three matchers. The cross-condition ordering and the full/mid/late separation are unchanged. The only material change is the mid-band rate ($15.5 \to 14.3$), from de-duplicating nested \texttt{sorry}/\texttt{i'm so sorry}; the small apparent transfer in the early and position-perturbing conditions ($0.3$) resolves to $0.0$ under boundary matching. The short marker \texttt{um} matched only literal \texttt{um} tokens (no \texttt{number}/\texttt{summary}/\texttt{assume} false positives) in this generation set.}
\label{tab:marker-sensitivity}
\end{table}

The reviewer-noted substring risk is therefore quantitatively immaterial to the reported ordering and dissociation pattern; the stricter matcher slightly \emph{reduces} the low-level marker counts in the non-target conditions, sharpening rather than weakening the separation.

\section{Hidden V$^\star$-gap vs.\ text-density scatter}
\label{app:hidden-vs-text}

\begin{figure}[H]
\centering
\includegraphics[width=0.85\linewidth]{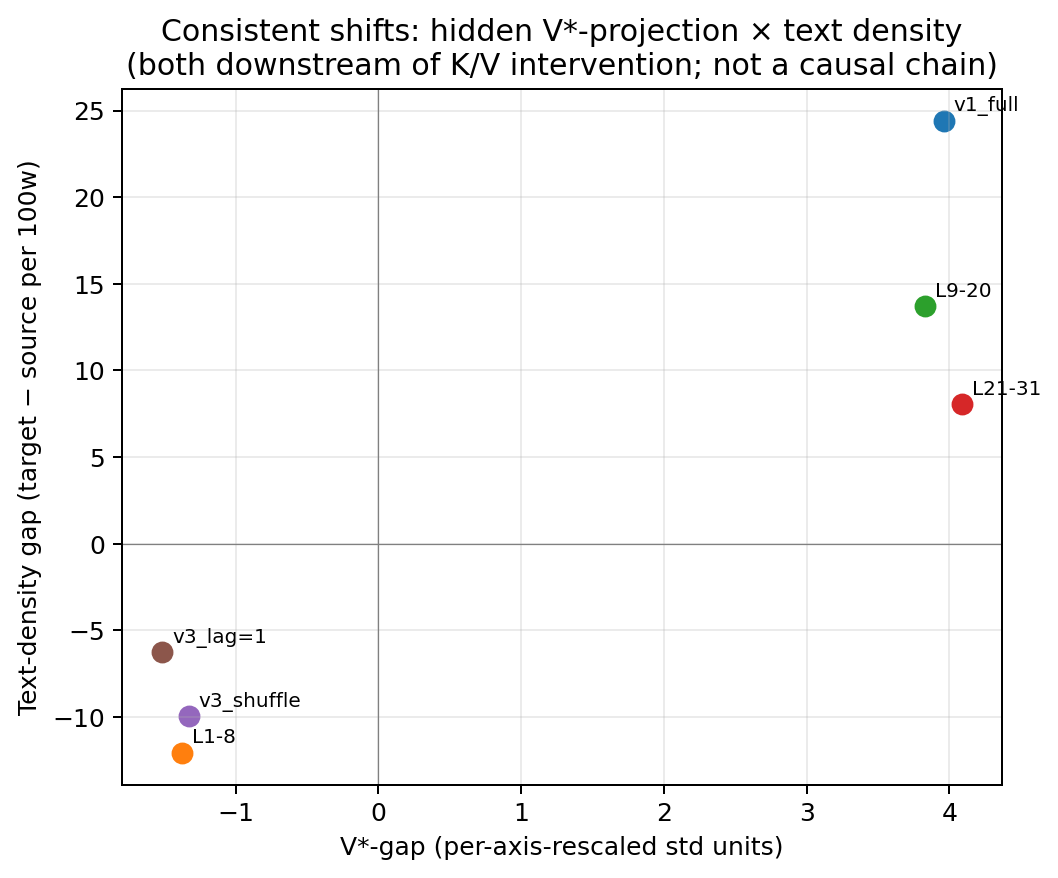}
\caption{Hidden-state V$^\star$-gap (per-axis-rescaled) vs.\ text-density gap, across the six conditions with L$28$ hidden-state measurements. Sign agreement holds throughout; magnitude relationships exhibit the dissociations characterized in \S\ref{sec:results-summary}.}
\label{fig:hidden-vs-text}
\end{figure}

\section{Full per-condition data table}
\label{app:full-data}

Table~\ref{tab:full-data} reports the complete per-condition summary for all $13$ K/V intervention configurations. ``Target/100w'' is target-persona-marker density per $100$ generated words; ``TTR'' is type--token ratio; ``V@L9--20'' is the V cosine gap at the L$9$--$20$ band; ``K@L9--20'' is the analogous K-space gap. Text means (Target/100w, Src/100w, TTR) are over $10$ samples per cell; K/V cosine quantities are from the single K/V dump per condition over $50$ generation steps, $8$ K/V heads, and the $12$ layers in L9--20 where applicable.

\begin{table}[H]
\centering
\small
\begin{tabular}{lrrrrr}
\toprule
Condition & Tgt/100w & Src/100w & TTR & V@L9-20 & K@L9-20 \\
\midrule
v1\_full              & $24.81$ & $0.40$  & $0.650$ & $+0.94$ & $+0.41$ \\
v2\_L1--8             & $0.29$  & $12.36$ & $0.846$ & $+0.02$ & $-0.03$ \\
v2\_L9--20            & $15.52$ & $1.82$  & $0.766$ & $+0.89$ & $+0.38$ \\
v2\_L21--31           & $10.47$ & $2.42$  & $0.393$ & $+0.04$ & $-0.03$ \\
v2\_L9--12            & $0.51$  & $7.75$  & $0.878$ & $+0.28$ & $+0.09$ \\
v2\_L13--16           & $5.61$  & $3.34$  & $0.843$ & $+0.36$ & $+0.12$ \\
v2\_L17--20           & $6.22$  & $12.09$ & $0.771$ & $+0.33$ & $+0.11$ \\
v3\_shuffle           & $0.29$  & $10.23$ & $0.875$ & $-0.07$ & $-0.08$ \\
v3\_lag$=1$           & $0.31$  & $6.57$  & $0.847$ & $-0.06$ & $-0.08$ \\
v3\_lag$=3$           & $0.28$  & $10.02$ & $0.841$ & $-0.07$ & $-0.08$ \\
v3\_lag$=5$           & $0.00$  & $10.41$ & $0.859$ & $-0.08$ & $-0.09$ \\
v4\_$\alpha{=}0.25$   & $0.57$  & $9.29$  & $0.851$ & $+0.21$ & $+0.03$ \\
v4\_$\alpha{=}0.50$   & $3.15$  & $3.39$  & $0.805$ & $+0.65$ & $+0.20$ \\
\bottomrule
\end{tabular}
\caption{Full per-condition data summary across the $13$-condition K/V intervention battery.}
\label{tab:full-data}
\end{table}

\section{K/V cosine analysis: supplementary measurements}
\label{app:kv-supp}

\paragraph{K-space locality.} Table~\ref{tab:kv-k-propagation} reports the K-space analog of Table~\ref{tab:kv-propagation}. K-space exhibits the same locality pattern as V-space, with substantially smaller magnitudes within intervened bands ($0.36$--$0.38$ vs.\ $0.84$--$0.91$). Off-band K-gaps are negative ($-0.03$ to $-0.11$).

\begin{table}[H]
\centering
\small
\begin{tabular}{lccc}
\toprule
Condition & K @ L1--8 & K @ L9--20 & K @ L21--31 \\
\midrule
v1\_full        & $0.36$         & $0.41$         & $0.39$         \\
v2\_L1--8       & $\mathbf{0.36}$ & $-0.03$        & $-0.10$        \\
v2\_L9--20      & $-0.09$        & $\mathbf{0.38}$ & $-0.03$        \\
v2\_L21--31     & $-0.11$        & $-0.03$        & $\mathbf{0.37}$ \\
\bottomrule
\end{tabular}
\caption{K-space cosine gap by measurement band $\times$ intervention condition. Bold entries indicate the intervened band.}
\label{tab:kv-k-propagation}
\end{table}

\paragraph{Position-perturbing and interpolation conditions.} v3\_lag$=1$ shows near-zero to slightly negative gaps across bands (V-gap $\in [-0.09, 0.00]$) and v3\_shuffle is similarly near zero ($|V| \leq 0.09$); both indicate disruption of source alignment without target injection. v4\_$\alpha = 0.50$ produces an L9--20 V-gap of $0.65$ (between L9--20-only's $0.89$ and no-intervention's $0.0$), with near-zero gaps at non-intervened bands.

\paragraph{Fine-grained L9--20 sub-band interventions.} Within the mid-layer band, V-gap measured at the full L9--20 band is intermediate for all three $4$-layer sub-bands: L9--12 $\to 0.28$; L13--16 $\to 0.36$; L17--20 $\to 0.33$. The corresponding target-marker densities differ markedly across sub-bands (L9--12 $\to 0.5$, L13--16 $\to 5.6$, L17--20 $\to 6.2$). The fine-grained ordering does not match the V-gap ordering monotonically, indicating that the sub-band sensitivity is not captured by a single V-gap scalar.

\section{Persistence of mid-layer K/V transfer to 150 tokens}
\label{app:t150}

The main K/V battery generates $50$ tokens (\S\ref{sec:setup}), which measures intervention
\emph{instantiation} rather than long-horizon persistence (\S\ref{sec:limitations}). As a direct
persistence check we re-ran v2\_L9--20 and v1\_full at $\texttt{max\_new\_tokens} = 150$ ($n = 3$
each), together with an un-intervened target-persona (Priya) baseline at $T = 150$ ($n = 5$) to
separate intervention effects from generation-length artifacts. Table~\ref{tab:t150} reports
target-marker density and TTR at cumulative word-truncation horizons.

\begin{table}[H]
\centering
\small
\begin{tabular}{lccc}
\toprule
Condition (target/100w / TTR) & @40w & @80w & @full \\
\midrule
v2\_L9--20 ($n{=}3$)     & $16.7 / 0.75$ & $12.1 / 0.63$ & $11.9 / 0.61$ \\
v1\_full ($n{=}3$)       & $28.3 / 0.57$ & $26.9 / 0.47$ & $26.9 / 0.47$ \\
Priya baseline ($n{=}5$) & $17.5 / 0.85$ & $12.5 / 0.72$ & $11.7 / 0.68$ \\
\bottomrule
\end{tabular}
\caption{Target-marker density and TTR (per $100$ words) at cumulative truncation horizons under
$T = 150$ generation. ``Priya baseline'' is un-intervened target-persona generation, included to
separate intervention effects from generation-length artifacts.}
\label{tab:t150}
\end{table}

Two observations. First, target-persona marker density \emph{persists}: mid-layer replacement holds
$16.7 \to 11.9$ markers per $100$ words across the horizon, closely tracking the un-intervened Priya
baseline ($17.5 \to 11.7$); full replacement holds $\approx 27$. Persona expression shows no
decay over the $150$-token window in these samples. Second, TTR declines with generation length ($0.75 \to 0.61$ for v2\_L9--20), but
the un-intervened baseline declines comparably ($0.85 \to 0.68$): the intervention sits a roughly
constant $\sim\!0.08$ TTR below the un-intervened baseline at every horizon, indicating a fixed
lexical-repetition cost rather than progressive degradation. These $n = 3$ results are a small-sample
persistence check rather than a full battery, but they indicate that mid-layer K/V persona transfer
persists past the $50$-token instantiation window measured in the main paper.

\section{Code release manifest}
\label{app:code}

The full reproducibility artifact (released at \texttt{[anonymized]} upon acceptance) includes:

\begin{itemize}
\item Source code for the K/V intervention pipeline (extending \texttt{transformers\_hooks.py} with \texttt{chat\_with\_kv\_replacement}, \texttt{chat\_with\_kv\_shuffle}, \texttt{chat\_with\_kv\_lag}, \texttt{chat\_with\_kv\_interpolation});
\item The $13$ intervention-configuration JSON files used for the experimental battery;
\item K/V dumps (per-step last-position K and V tensors at full head dimension, fp$32$; per-step logits at fp$16$; sampled tokens) for one sample per condition, source-baseline, and target-baseline---total approximately $1$~GB;
\item Analysis scripts: text-density (\texttt{03g\_kv\_hack\_analysis.py}), hidden V$^\star$-projection (\texttt{03j\_kv\_hidden\_v\_star\_proj.py}), K/V cosine and logit-shift bridge (\texttt{03i\_kv\_cosine\_analysis.py}), main-figure rendering (\texttt{03h\_kv\_main\_figure.py});
\item The V$^\star$ basis NPZ and persona-centroid CSV (small files, $<5$~MB) inherited from prior work, released alongside;
\item Closed-loop V$^\star$ state-feedback hook (\texttt{closed\_loop\_v\_star\_hook.py}), context-window ablation script, and open-loop pilot script for the activation-channel experiments in Appendix~\ref{app:activation-channel};
\item README and reproducibility-instruction files documenting expected directory structure, dependencies (\texttt{transformers}, \texttt{torch}, \texttt{matplotlib}, \texttt{numpy}), and command-line invocations.
\end{itemize}

\end{document}